%% file: main.tex
\documentclass{article} 
\usepackage{iclr2026_conference,times}
\usepackage{multirow}
\usepackage{amsmath}
\usepackage{amssymb}
\usepackage{graphicx}
\usepackage{tikz-cd}

\input{math_commands.tex}

\usepackage{hyperref}
\usepackage{url}
\usepackage{booktabs}       
\usepackage{amsfonts}       
\usepackage{nicefrac}       
\usepackage{microtype}      
\usepackage{xcolor}         
\usepackage{bm}

\title{From Cheap Geometry to Expensive Physics: Elevating Neural Operators via Latent Shape Pretraining}

\author{
  Zhizhou Zhang\thanks{Correspondence to: \texttt{zhizhou.zhang@cn.bosch.com}}\quad Youjia Wu\quad Kaixuan Zhang\quad Yanjia Wang \\
  Corporate Research \\
  Bosch (China) Investment Co., Ltd. \\
  \texttt{\{zhizhou.zhang, youjia.wu, kaixuan.zhang\}@cn.bosch.com} \\
}

\iclrfinalcopy 
\begin{document}

\maketitle

\input{base/0_abstract}
\input{base/1_introduction}
\input{base/2_setting}
\input{base/3_method}
\input{base/4_experiments}
\input{base/6_conclusion}


\newpage

\section*{Reproducibility statement}

Detailed descriptions of the experimental setup, task definitions, and evaluation metrics are provided in section~\ref{sec:exp} and Appendix~\ref{appendix:dataset}.

\bibliography{ref}
\bibliographystyle{iclr2026_conference}

\input{base/appendix}

\end{document}

%% file: math_commands.tex
\usepackage{amsmath,amsfonts,bm}

\def\eqref#1{equation~\ref{#1}}

\def\1{\bm{1}}

\DeclareMathAlphabet{\mathsfit}{\encodingdefault}{\sfdefault}{m}{sl}
\SetMathAlphabet{\mathsfit}{bold}{\encodingdefault}{\sfdefault}{bx}{n}



%% file: base/0_abstract.tex
\begin{abstract}

Industrial design evaluation often relies on high-fidelity simulations of governing partial differential equations (PDEs). While accurate, these simulations are computationally expensive, making dense exploration of design spaces impractical. Operator learning has emerged as a promising approach to accelerate PDE solution prediction; however, its effectiveness is often limited by the scarcity of labeled physics-based data. At the same time, large numbers of geometry-only candidate designs are readily available but remain largely untapped. 
We propose a two-stage framework to better exploit this abundant, physics-agnostic resource and improve supervised operator learning under limited labeled data. In \textbf{Stage 1}, we pretrain an autoencoder on a geometry reconstruction task to learn an expressive latent representation without PDE labels. In \textbf{Stage 2}, the neural operator is trained in a standard supervised manner to predict PDE solutions, using the pretrained latent embeddings as inputs instead of raw point clouds. Transformer-based architectures are adopted for both the autoencoder and the neural operator to handle point cloud data and integrate both stages seamlessly. Across four PDE datasets and three state-of-the-art transformer-based neural operators, our approach consistently improves prediction accuracy compared to models trained directly on raw point cloud inputs. These results demonstrate that representations from physics-agnostic pretraining provide a powerful foundation for data-efficient operator learning.

\end{abstract}

%% file: base/1_introduction.tex
\section{Introduction}
\label{sec: intro}

Partial differential equations (PDEs) are fundamental descriptors for various physical and engineering systems~\citep{guenther1996partial}. Traditionally, PDEs are solved numerically by various discretization methods such as finite element, finite difference, finite volume, etc.~\citep{reddy1993introduction,perrone1975general} which offer high accuracy at huge computation cost~\citep{greenspan1955methods}. This costly simulation process can be significantly accelerated by training surrogate neural operators to learn mappings from input functions to PDE solutions. As the pioneer neural operator work, DeepONet adopted MLP structures and established theoretical foundations for operator learning~\citep{lu2021learning}. Kernel-based operators~\citep{li2020fourier,li2020neural} are then proposed to better handle resolution invariance. Recent studies focus on transformer-based neural operators~\citep{cao2021choose,hao2023gnot,wu2024transolver} to handle complex irregular geometry inputs and further improve computation efficiency.

Despite architectural progress, the accuracy of neural operators still depends heavily on labeled PDE solutions produced by expensive simulations. To mitigate data scarcity, several works adopt masked reconstruction pretraining on neural operator inputs~\citep{chen2024data,rahman2024pretraining}. This is motivated by the widely adopted pretraining techniques in computer vision and natural language processing~\citep{he2022masked,devlin2019bert}, but is only applicable to problems with fixed grid structures. \cite{deng2024geometry} couples neural operators with a ULIP-2 pretrained Point-BERT, which assumes surface point cloud inputs. \cite{cheng2024reference} proposes to predict deviation from a reference solution, requiring grid topology correspondence across data samples. These methods are effective within their constraints, but their applicability to general PDE problems with versatile input formats remains limited.

A complementary direction explores supervised pretraining on large scale PDE solutions with related underlying physics. \cite{serrano2024aroma} trains a latent space via physical field reconstruction and perform latent rollouts to stabilize dynamics. \cite{hao2024dpot} and \cite{mccabe2024multiple} train foundation spatiotemporal neural operator on large scale datasets covering different variants of computation fluid dynamic (CFD) PDEs, showing enhanced accuracy when finetuned on small fluid datasets. In industrial practice, however, PDE solution labels are scarce and assembling large physics datasets is rarely feasible. By contrast, geometry only data (mesh without solver labels) are plentiful yet underused. This physics-agnostic resource offers an untapped path to improve operator learning under label scarcity.

\begin{figure}[!ht]
    \centering
    \includegraphics[width=0.9\linewidth]{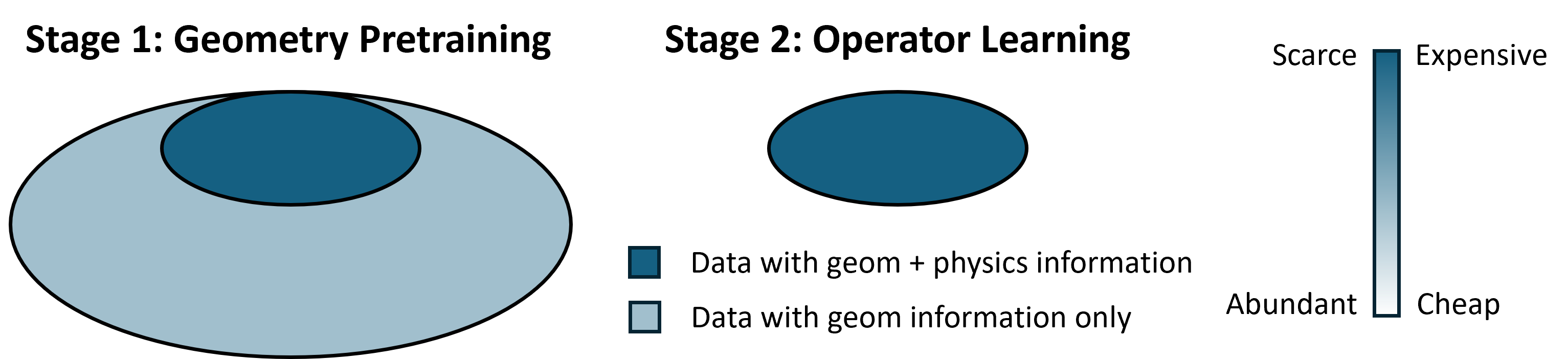}
    \caption{Illustration of the proposed two-stage training framework. Stage 1 leverages the abundant geometry data for pretraining. Stage 2 learns PDE solutions on the scarce physics data.}
    \label{fig:pre_illustration}
\end{figure}

To exploit the abundant geometry information while respecting the limited PDE solution labels, we propose the replace the standard end-to-end supervised recipe with a two-stage training pipeline as shown in Fig.~\ref{fig:pre_illustration}. Stage 1 pretrains a point cloud variational autoencoder (VAE) to extract latent geometry representation using a large point cloud/mesh dataset. A physics-agnostic proxy task is introduced as the reconstruction objective for VAE training. In stage 2, neural operators are trained to predict PDE solutions from the stage 1 latents rather than raw point clouds. As stage 1 is exposed to far more geometries than labeled PDE solutions, the pretrained encoder can regularize and compensate for point cloud inputs (which are downsampled realizations of geometries), yielding more informative representations and improving data efficiency on scarce PDE datasets. The main contribution of our framework are summarized as follows:

\begin{itemize}
    \item We propose a novel two-stage training framework for operator learning tasks. Stage 1 performs pretraining on unlabeled point clouds via a proxy task to learn latent geometry representation. Stage 2 trains neural operators to predict PDE solutions from these latents. This training scheme allows full exploitation of geometry-only data without PDE labels.
    \item We introduce occupancy reconstruction as the objective of stage 1 pretraining, allowing physics-agnostic self-supervised representation learning on irregular geometries realized as point clouds. The selection of proxy task is flexible, which can be swapped for signed distance fields, shortest vector fields, and other choices.
    \item We examine the two-stage training framework across four PDE problems and three transformer-based neural operators. Experiments show consistent performance gain when compared with direct single stage supervised training.
\end{itemize}

%% file: base/2_setting.tex
\section{Preliminaries}

In this section, we define the problem and data settings. Then we briefly summarize related works on operator learning and pretraining.

\subsection{Problem Setup}
\label{sec:setting}
We consider PDE problems defined in the domain $\Omega\subset \mathbb{R}^d$ to be approximated with a learned operator $\mathcal{F}:A\rightarrow U$ where $A$ represents the input function space and $U$ represents the solution function space of the physics over $\Omega$. In general scenarios, $A$ may consist of various types of information such as object geometries (material distribution), boundary conditions, source functions, etc. For simplicity, the input function space is assumed to contain only object geometry information, and the solution function is assumed to be scalar valued in this work. A realization of the geometry information $a_k\in A$ is represented as a point cloud (nodes of a mesh) $a_k=\{x^i_k\}^N_{i=1}=\mathbf{X}_k\in \mathbb{R}^{N\times d}$ with $N$ points in domain $\Omega$. And the physics solution function $u_k\in U$ is realized as $\{(y^i_k,u^i_k)\}^{N'}_{i=1}$ with $u^i_k=u_k(y^i_k)$, where $\{y^i_k\in\Omega\}^{N'}_{i=1}$ is a set of query points.

In this work, we consider two qualitatively different portions $D$ and $D'$ from each dataset:
\begin{equation}
\label{eq:dataset}
\begin{split}
D= \lbrace (a_k,u_k) \rbrace ^ {|D|} _ {k=1}, \quad D'= \lbrace (a_j) \rbrace ^ {|D'|} _ {j=1}, \quad |D'| \gg |D|
\end{split}
\end{equation}
Here, $D$ is the physics portion with ground truth fields produced by a numerical solver, while $D'$ is the geometry-only portion that is abundant. This is highly related to many industrial scenarios where geometry designs are easy to generate, but physical field labels are limited. Motivated by such scenarios, we aim to leverage $D'$ to improve prediction accuracy on $D$, thereby alleviating the persistent data-scarcity bottleneck in operator learning.

\subsection{Related Work}
\label{sec:related}
\paragraph{Neural Operators.} Operator learning aims to establish a mapping from input functions to solution functions. DeepONet established theoretical foundations for operator learning~\citep{lu2021learning}, followed by two important milestone: Graph Neural Operator (GNO)~\citep{li2020neural} and Fourier Neural Operator (FNO)~\citep{li2020fourier}. A set of follow-up works further extended GNO and FNO to irregular geometries~\citep{li2023geometry,li2023fourier} and improved computation efficiency~\citep{guibas2021adaptive,tran2021factorized}. With the rise of attention mechanism, architectural design shifted toward transformers~\citep{cao2021choose,li2022transformer}, where spatial locations and field values are tokenized as query, key, and value sequences. General Neural Operator Transformer (GNOT)~\cite{hao2023gnot} introduced heterogeneous cross-attention to handle versatile input functions and gating mechanism to adjust network focus. Universal Physics Transformers (UPT)~\cite{alkin2024universal} adopts an encoder-decoder architecture to roll out physics evolution in the latent space, which an end-to-end training process. Transolver \cite{wu2024transolver} proposed the usage of physics attention to efficiently pathify point based input tokens into learnable slices, which are projected back to query coordinates at the ouput layer. Latent Neural Operator (LNO)~\cite{wang2024latent} proposed the invertible physics-cross-attention to efficiently perform compute attention on latent tokens. Overall, transformer-based neural operators have achieved state of the art in accuracy, offer greater input flexibility, and exhibit strong parallelization potential.

\paragraph{Operator Pretraining.} Self-supervised pretraining on large dataset has proved effective in natural language processing tasks~\cite{devlin2019bert,raffel2020exploring}, computer vision tasks~\cite{he2022masked,dosovitskiy2020image}, and some scientific tasks~\cite{zhou2023uni,nguyen2023climax}. Following a similar practice, \cite{chen2024data} and \cite{rahman2024pretraining} pretrains neural operators by masking and reconstructing data inputs on fixed grid structures. 3D-GeoCA~\citep{deng2024geometry} attempts to directly guide neural operators via Point-BERT pretrained on ULIP-2~\citep{xue2024ulip} for processing surface grids. These works demonstrate the potential of unsupervised pretraining in operator learning, but requires specific input formats. AROMA~\citep{serrano2024aroma} proposes to pretrain a latent space via physics reconstruction to stabilize long horizon rollouts for dynamic problems. MPP~\citep{mccabe2024multiple} and DPOT~\citep{hao2024dpot} demonstrate effectiveness of autoregressive pretraining on spatiotemporal fields across datasets that share similar underlying fluid dynamics. However, these studies focus primarily on squeezing more values from PDE solutions while leaving geometry data untapped. This highlights a key gap: developing physics-agnostic pretraining strategies for neural operators on general irregular geometries.




%% file: base/3_method.tex
\section{Methods}
\label{sec:methods}

\begin{figure}[!ht]
    \centering
    \includegraphics[width=0.96\linewidth]{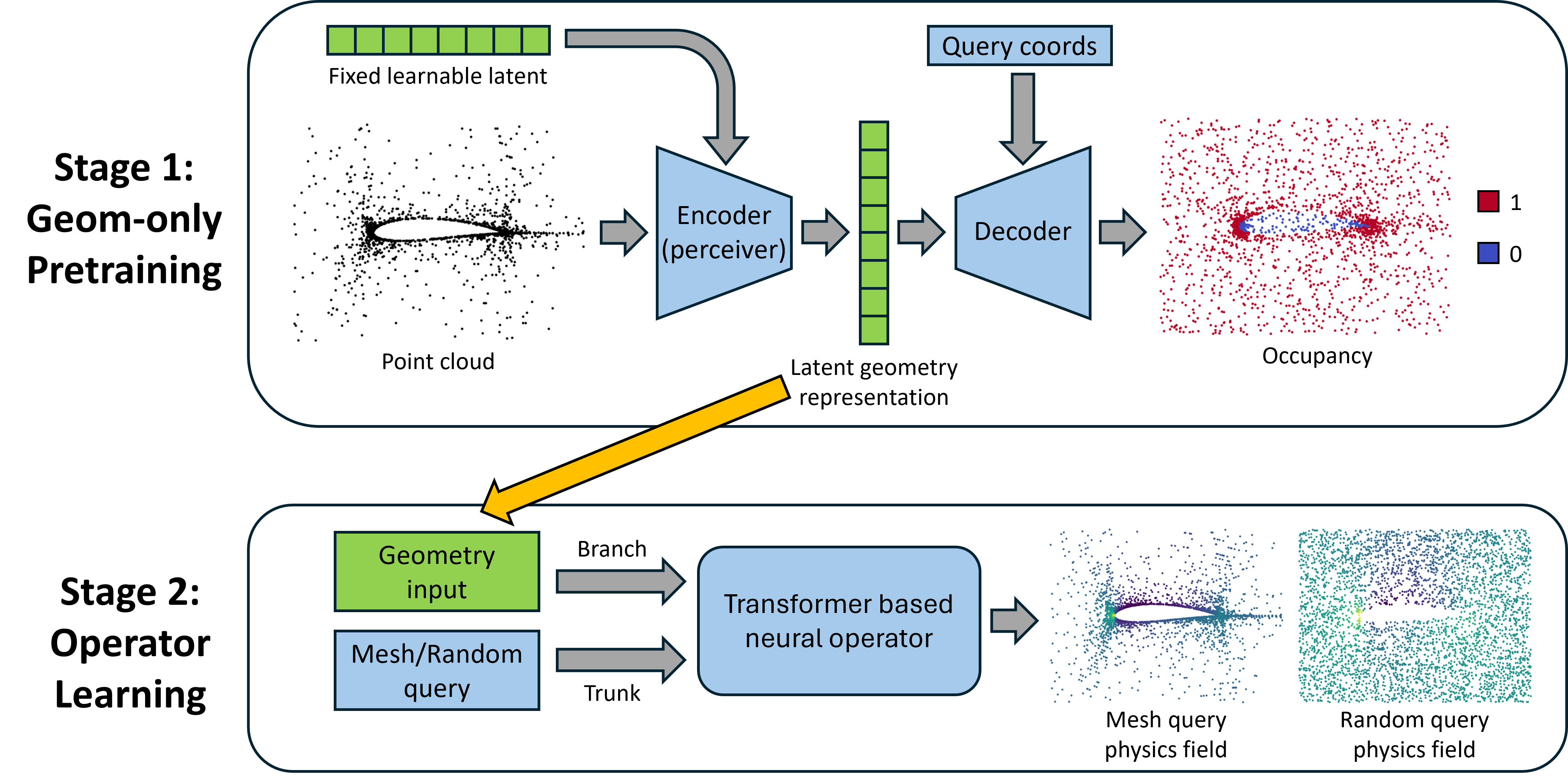}
    \caption{An illustration of our two-stage training framework. In stage 1, an encoder is pretrained on rich geometry data using a physics-agnostic proxy task, for which we choose occupancy field. The encoded latent representation is then integrated to transformer-based neural operators for stage 2 training on scarce PDE solution labels. The performance is tested on two query sampling strategies: mesh-based query, and uniform random queries.}
    \label{fig:main_illustration}
\end{figure}

We propose to split the entire training process into two stages as seen in Fig.~\ref{fig:main_illustration}. Stage 1 performs pretraining on physics-agnostic $(a,o)$ data pairs from $D\cup D'$ to learn a better representation from input point clouds $a$, where $o\in O$ represents a proxy task $\mathcal{G}:A\rightarrow O$ that can be obtained from $a$ at negligible computation cost. The two datasets are therefore augmented to $D=\{(a,o,u)\}^{|D|}$ and $D'=\{(a,o)\}^{|D'|}$. Stage 2 leverages the representation from stage 1 and follows common neural operator training practice on $D$ to predict $u$ from $a$.


\subsection{Physics-Agnostic Geometry Pretraining via Occupancy Prediction}
\label{sec:occupancy}
To obtain a meaningful representation from stage 1 pretraining, a proper proxy task should possess the following three features: \textbf{(1) Computation efficiency.} It should be broadly applicable and computable directly from the mesh/point cloud input $a$ at low computation cost. \textbf{(2) Consistency with operator learning.} It should stay under the umbrella of operator learning tasks to bridge naturally towards physical field prediction. \textbf{(3) Sampling invariance.} It should treat different point cloud realizations of the same underlying geometry equivalently as discussed in Appendix~\ref{appendix:principle}. Ideally, stage 1 pretraining would encode regions of high expected physical significance to better inform the operator training in stage 2. However, in a fully general setting without explicit physics supervision, it is difficult to design an inductive bias that consistently improves performance across all problems. Taking all these factors into consideration, we pick occupancy field~\citep{mescheder2019occupancy} $o\in O$ as the proxy task for the main experiments, which is realized (similar to $u$) by its values on a set of query points $\{(z^i_k,o^i_k)\}^{N''}_{i=1}$ with $o_k^i=o_k(z^i_k)\in \{0,1\}$, where $o(z)=1$ indicates that $z$ is inside an object and $o(z)=0$ vice versa. Note that our framework can be easily extended to other proxy tasks. In section~\ref{subsec: ablation} we also explore signed distance fields (SDF) and shortest vectors (SV)~\citep{jessica2023finite} as alternative proxy tasks, while reserving occupancy fields for the main experiments.

To extract geometry representation from the occupancy proxy task, we follow the practice in~\cite{zhang20233dshape2vecset} to train a point cloud variational autoencoder (VAE) in stage 1 as shown in Fig.~\ref{fig:main_illustration}. The encoder $\mathcal{E}$ adopts a perceiver architecture for aggregating the point cloud information into a fixed length vector, which is then projected into a probabilistic latent space via $MLP_\mu:\mathbb{R}^C\rightarrow\mathbb{R}^{C_0}$ and $MLP_\sigma:\mathbb{R}^C\rightarrow\mathbb{R}^{C_0}$:
\begin{equation}
\label{eq:encoder}
\begin{split}
    m^k &= CrossAttn(\mathbf{L},PosEmb(a_k=\mathbf{X}_k))\\
    (h_\mu)_k^i &= MLP_{\mu}(m^k_i)_{i\in[1,2,...M]}\\
    (h_\sigma)_k^i &= MLP_{\sigma}(m^k_i)_{i\in[1,2,...M]}
\end{split}
\end{equation}
where $\mathbf{X}\in\mathbb{R}^{N\times d}$ is the $d$ dimensional point cloud of $N$ points, $PosEmb: \mathbb{R}^d\rightarrow\mathbb{R}^C$ is a positional embedding neural network, $\mathbf{L}\in\mathbb{R}^{M\times C}$ is a set of learnable tokens with fixed number $M$. The latent representation $z\in\mathbb{R}^{M\times C_0}$ is then sampled as:
\begin{equation}
\label{eq:sample_latent}
    h_k^i = (h_\mu)_k^i+(h_\sigma)_k^i\cdot\epsilon
\end{equation}
where $\epsilon\sim\mathcal{N}(0,1)$. The decoder then predicts the occupancy value of a set of query points to reconstruct the geometry $o_k(z_i)=\mathcal{D}(h_k)(z_i)$, so that the latent representation resides in function space (Appendix~\ref{appendix:principle}). The training objective for the point cloud VAE takes the following form:
\begin{equation}
\label{eq:vae_loss}
\min_{\phi,\eta}\frac{1}{|D\cup D'|}\sum_{k=1}^{|D\cup D'|}(\mathbb{E}_{z\sim p}\mathrm{BCE}(\tilde{\mathcal{D}}_\eta(\tilde{\mathcal{E}}_\phi(a_k))(z),o_k(z))+\lambda\cdot\mathrm{KL}(\mathcal{N}(h_\mu,h_\sigma) \,\|\, \mathcal{N}(0,1)))
\end{equation}


where $\mathrm{BCE}(\cdot,\cdot)$ stands for the binary cross entropy loss, and $\lambda$ is the KL weight. The sampling strategy $z\sim p$ for occupancy field consists of two parts: a uniform distribution over the computation domain $\mathcal{U}(\Omega)$, and perturbed point cloud $z^i=x^i+\varepsilon^i$ where $\varepsilon\sim\mathcal{N}(0,\epsilon I)$. Physics datasets are oftentimes sparse as they are computationally expensive to collect. However, the geometry dataset can be dense ($|D'|\gg |D|$) as the computation cost for generating new geometries is negligible.

\subsection{Stage 2 Operator Learning}

Instead of directly taking $a$ as inputs, neural operators $\mathcal{F}$ take advantage of the geometry representation learned from stage 1 pretraining (Fig.~\ref{fig:main_illustration}) in the form of $\tilde{\mathcal{F}}_\theta (\tilde{\mathcal{E}}_\phi(a_k))=u_k$, where $\theta$ is a set of trainable parameters. Notice that the latent tokens produced by the pretrained encoder $\tilde{\mathcal{E}}_\phi$ can be smoothly integrated to transformer based neural operators discussed in section~\ref{sec:related}. The training goal is to minimize the relative L2 error where $\hat{u}_k$ denotes the normalized ground truth physical field.

\begin{equation}
\label{eq:metric}
\min_{\theta}\frac{1}{|D|}\sum_{k=1}^{|D|}\frac{\sqrt{\sum_{i=1}^{N'}(\tilde{\mathcal{F}}_\theta^i(\tilde{\mathcal{E}}_\phi(a_k))-\hat{u}_k^i)^2}}{\sqrt{\sum_{i=1}^{N'}(\hat{u}_k^i)^2}}
\end{equation}

In stage 2 training, we only update parameters $\theta$ while keeping the encoder $\tilde{\mathcal{E}}_\phi$ frozen. In this work, we consider two types of different query strategies for $u$ that are widely used in various operator learning studies: directly querying the physical field on the point cloud (mesh) $\{y^i_k\}=\{x^i_k\}^N_{i=1}$, and querying the physical field according to a set of uniformly sampled random points $\{y^i_k\sim\mathcal{U}(\Omega)\}^{N'}_{i=1}$.

%% file: base/4_experiments.tex
\section{Experiments}
\label{sec:exp}

We evaluate the proposed two-stage training framework on four different datasets as described in section~\ref{subsec:datasets}. The quantitative evaluation of our method is presented in sections~\ref{subsec: main_results} and~\ref{subsec: ablation}, showing that a learned latent geometry representation from physics-agnostic pretraining can enhance neural operator's understanding of physics fields.

\subsection{Datasets}
\label{subsec:datasets}

To examine our method, we modified the Elasticity~\citep{li2023fourier} and AirfRans~\citep{bonnet2022airfrans} datasets and built two new datasets to generate occupancy fields for stage 1 physics-agnostic pretraining, and physical field for  stage 2 operator learning. The geometries in $D'$ are sampled from the same (or similar) distribution as the geometries in $D$. We list the general statistics (Table~\ref{table:dataset}) and the generation process of the datasets as the following. More details on datasets can be found in Appendix~\ref{appendix:dataset}.

\paragraph{Stress.} We mimick the practice of the Elasticity dataset~\citep{li2023fourier} to generate occupancy fields for stage 1 pretraining, and Von Mises stress fields for stage 2 operator learning.

\paragraph{AirfRans (near).} We directly adopt the pressure field data from AirfRans~\citep{bonnet2022airfrans} and truncate the domain ($[-0.6,1.6]\times[-0.5,0.5]$) to focus only on the near volume of the airfoils. We then follow the instruction in \citep{bonnet2022airfrans} to generate new geometries via OpenFOAM and compute occupancy fields without running the CFD simulations.

\paragraph{Inductor (3D).} We solve the Maxwell's equations using Comsol for a parameterized 3D iron core subject to the current excitation in a copper coil. The norm of the magnetic flux density field is exported as the physics field to be predicted. Occupancy fields are also generated for the parameterized iron cores.

\paragraph{Electrostatics.} We solve the electrostatics Poisson equation in a domain that contains two materials with distinct permeability values. The two material domains are separated by a curved interface generated from a third order spline interpolation of 10 random points. We then compute the occupancy fields for stage 1 pretraining, and electric potential fields for stage 2 operator learning.

\begin{table}[!ht]
  \caption{Statistics of the four datasets for the main experiments. The second row shows the size of physics data labeled with PDE solutions $u$, while the third row shows the total data size including physics and geometry-only data.}
  \label{table:dataset}
  \centering
  \renewcommand{\arraystretch}{1.05}
  \begin{tabular}{ccccc}
    \hline\hline
    Statistics & Stress & AirfRans (near) & Inductor (3D) & Electrostatics \\
    \hline
    Dimension & 2D & 2D & 3D & 2D \\
    $D$  (train+test) & 1900+200 & 800+200 & 900+300 & 1800+200 \\
    $D\cup D'$   (train+test) & 7600+500 & 2400+600 & 3600+300 & 7600+400 \\
    \hline\hline
  \end{tabular}
\end{table}

\begin{figure}[ht]
    \centering
    \includegraphics[width=0.98\linewidth]{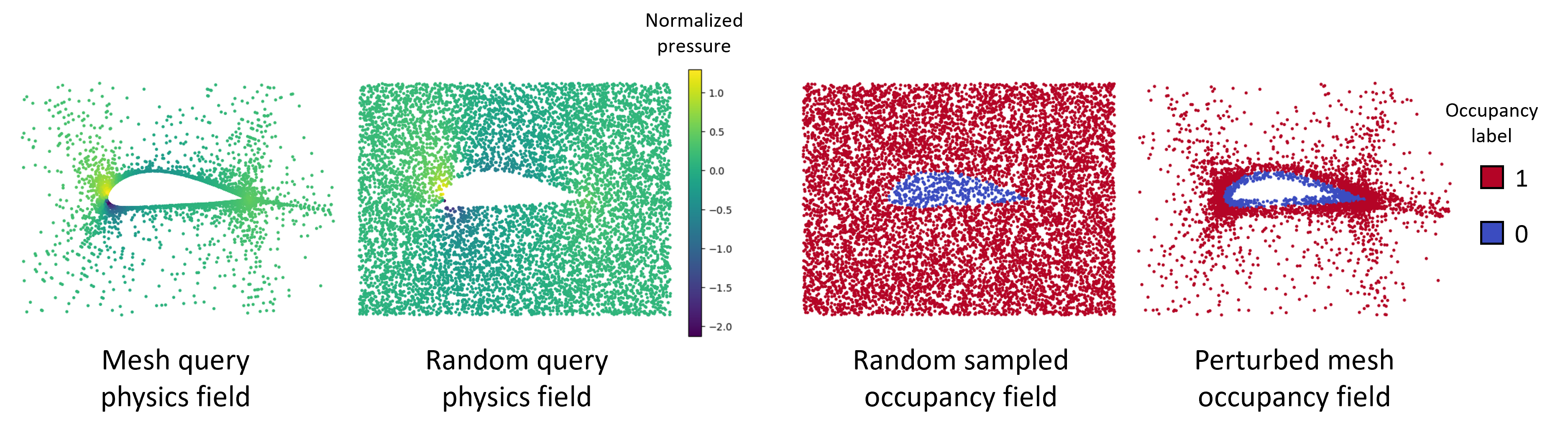}
    \caption{Visualization of a sample data point from the near volume AirfRans dataset. All of the datasets are preprocessed so that each data point consists of four types of information: physics field queried on the mesh point cloud, physics field queried on randomly sampled points, occupancy field queried on randomly sampled points, and occupancy field queried on the mesh point cloud perturbed by random displacements. \textbf{The geometry data $D'$ are labeled only with occupancy fields.}}
    \label{fig:data_illustration}
\end{figure}

All four datasets follow the format of the sample data point visualized in Figure~\ref{fig:data_illustration}. Stage 1 pretraining is performed on a combination of the two occupancy field sampling strategy on $D\cup D'$ as discussed in section~\ref{sec:occupancy}. In stage 2, we perform operator learning on $D$ and report the physics field prediction accuracy on random uniform queries and mesh-based queries separately. We also concatenate the occupancy value $o^i_k$ to the query coordinates $y^i_k$ for training neural operators, so that the solution function takes the form $\{([y^i_k,o_k^i],u_k^i)\}^{N'}_{i=1}$. Note that for mesh-based queries where $\{y^i_k\}=\{x^i_k\}^N_{i=1}$, occupancy value is always 1.

\subsection{Models and training.}

We benchmark three transformer-based neural operators GNOT~\citep{hao2023gnot}, Transolver~\citep{wu2024transolver}, LNO~\citep{wang2024latent} by training each under the standard operator learning setting and under our proposed two-stage framework for performance comparison. The models are adjusted to have approximately the same number of learnable parameters. \textbf{Notice that we intentionally avoid exhaustive tuning of model or training hyperparameters, as our focus is on evaluating the effectiveness of the two-stage training strategy. Therefore, the reported performance should not be interpreted as a definitive comparison of model superiority.}

\paragraph{Point cloud VAE.} We adopt the point cloud VAE architecture in~\cite{zhang20233dshape2vecset} and train the model using Eq~\ref{eq:vae_loss}. We set the embedding dimension $C$ to 256, latent dimension $C_0$ to 32, learnable token number $M$ to 256, and depth of self-attention layers on latent space to 6.

\paragraph{GNOT.} We adopt the GNOT model proposed in~\citep{hao2023gnot}, with a hidden size of 128, head number of 4, and layer number of 4 (each layer includes a linear self-attention and a linear cross-attention block). The gating strategy is excluded from the model to keep the comparison concise. The latent geometry representation can be fed directly into the GNOT branch net without modification.

\paragraph{Transolver.} We adopt the Transolver model proposed in~\cite{wu2024transolver}, with a hidden size of 128, head number of 4, slice number of 32, and 9 layers of physics-attention. The ``get grid" layer is excluded from the model to keep the comparison concise. As the vanilla Transolver model assumes coupled query positions and point cloud/function observation positions, we replace the first physics-attention layer of Transolver with a linear cross-attention layer from GNOT to allow for latent geometry representation inputs.

\paragraph{LNO.} We adopt the LNO model proposed in~\citep{wang2024latent}, with a hidden size of 128, head number of 4, and 8 attention layers. The vanilla LNO assumes shared parameters for encoding position information of the trunk input and branch input. Instead, we use two distinct MLPs to allow for latent geometry representation inputs to the branch net.

\paragraph{Training strategy.} In each VAE training iteration, 2048 points are sampled from the mesh to explicitly represent the geometry. 1024 points are sampled from the random occupancy field, and another 1024 points are sampled from the perturbed mesh occupancy field (illustrated in Figure~\ref{fig:data_illustration}) to estimate the geometry reconstruction error. We run 400 epochs using AdamW, with a half-cycle cosine scheduler of learning rate $1\times10^{-3}$ and minimum learning rate $1\times10^{-6}$. All training processes of VAEs are completely agnostic of physics.

For all datasets, the first 100 physics data are used to normalize the physics field. We then compute relative L2 error (Eq~\ref{eq:metric}) on the normalized physics field for training loss and evaluation metric. \textbf{Notice that the relative L2 error can appear much smaller on unnormalized datasets with large absolute values in the mean.} We run 200 epochs of training for experiments on neural operators, using the same optimizer and scheduler strategy as VAEs. In each operator training iteration, 2048 points are sampled from the mesh to explicitly represent the geometry, 4096 points are sampled from either the mesh query or random query physics field to estimate the operator prediction error. More details of GPU usage and compute resources are included in Appendix~\ref{appendix: computation}.

\subsection{Main Results}
\label{subsec: main_results}

We compare the performance of transformer-based neural operators between taking point clouds as inputs and taking pretrained latent representation from stage 1 as inputs. Table~\ref{table:main_exp} shows the operator prediction error under mesh query and random query, with and without the encoder from stage 1 pretraining. The VAEs and neural operators in the main experiments are trained with datasets listed in Table~\ref{table:dataset}, with a fixed KL weight of $0.001$.

\begin{table}[!ht]
  \caption{Comparison of prediction errors on the four datasets. We report the mean error, with the standard deviation from three independent trials shown in parentheses. The two-stage training framework is tested on three transformer-based neural operators and compared with results from standard operator learning. We use bolded font to highlight the results when stage 1 pretraining enhances the performance of stage 2 operator learning.}
  \label{table:main_exp}
  \centering
  \renewcommand{\arraystretch}{1.05}
  \begin{tabular}{cccccccc}
    \hline\hline
    \multirow{2}{*}{Dataset} & \multirow{2}{*}{Query} & \multicolumn{6}{c}{Relative L2 on normalized data ($\times 10^{-2}$)} \\
    \cline{3-8}
       &      & GNOT & G+VAE & Trans & T+VAE & LNO & L+VAE \\
    \hline\hline
    \multicolumn{2}{c}{Number of parameters} & \multicolumn{2}{c}{1.7-1.8 M} & \multicolumn{2}{c}{1.7-1.8 M} & \multicolumn{2}{c}{1.8-1.9 M}\\
    \hline
    \multirow{2}{*}{Stress} & Mesh & \makebox[3.5em][r]{9.8(0.2)} & \makebox[3.5em][r]{\textbf{9.0(0.1)}} & \makebox[3.5em][r]{11.5(0.2)} & \makebox[3.5em][r]{11.2(0.2)} & \makebox[3.5em][r]{26.5(1.3)} & \makebox[3.5em][r]{\textbf{13.6(0.6)}}  \\
    & Random & \makebox[3.5em][r]{10.3(0.2)} & \makebox[3.5em][r]{\textbf{8.3(0.1)}} & \makebox[3.5em][r]{11.5(0.6)} & \makebox[3.5em][r]{\textbf{9.7(0.2)}} & \makebox[3.5em][r]{20.0(0.2)} & \makebox[3.5em][r]{\textbf{11.6(0.8)}} \\
    \hline
    \multirow{2}{*}{AirfR (near)} & Mesh & \makebox[3.5em][r]{6.8(0.2)} & \makebox[3.5em][r]{\textbf{5.6(0.1)}} & \makebox[3.5em][r]{13.4(0.3)} & \makebox[3.5em][r]{12.7(0.3)} & \makebox[3.5em][r]{27.4(1.3)} & \makebox[3.5em][r]{27.1(0.2)}    \\
    & Random & \makebox[3.5em][r]{7.8(0.2)} & \makebox[3.5em][r]{\textbf{5.9(0.1)}} & \makebox[3.5em][r]{15.0(0.2)} & \makebox[3.5em][r]{\textbf{10.8(0.1)}} & \makebox[3.5em][r]{25.3(1.8)} & \makebox[3.5em][r]{\textbf{10.0(0.2)}} \\
    \hline
    \multirow{2}{*}{Inductor (3D)} & Mesh & \makebox[3.5em][r]{7.0(0.1)} & \makebox[3.5em][r]{7.1(0.1)} & \makebox[3.5em][r]{11.4(0.5)} & \makebox[3.5em][r]{\textbf{8.4(0.2)}} & \makebox[3.5em][r]{24.9(1.4)} & \makebox[3.5em][r]{\textbf{9.2(0.6)}}  \\
    & Random & \makebox[3.5em][r]{12.5(0.2)} & \makebox[3.5em][r]{\textbf{11.8(0.2)}} & \makebox[3.5em][r]{16.8(0.6)} & \makebox[3.5em][r]{\textbf{13.2(0.2)}} & \makebox[3.5em][r]{20.3(1.0)} & \makebox[3.5em][r]{\textbf{13.0(0.2)}} \\
    \hline
    \multirow{2}{*}{Electrostatics} & Mesh & \makebox[3.5em][r]{4.2(0.1)} & \makebox[3.5em][r]{\textbf{3.3(0.1)}} & \makebox[3.5em][r]{5.0(0.1)} & \makebox[3.5em][r]{\textbf{3.8(0.1)}} & \makebox[3.5em][r]{13.5(0.4)} & \makebox[3.5em][r]{\textbf{4.6(0.2)}}  \\
    & Random & \makebox[3.5em][r]{4.6(0.2)} & \makebox[3.5em][r]{\textbf{3.4(0.0)}} & \makebox[3.5em][r]{5.6(0.1)} & \makebox[3.5em][r]{\textbf{3.9(0.1)}} & \makebox[3.5em][r]{13.5(0.2)} & \makebox[3.5em][r]{\textbf{4.7(0.1)}} \\
    \hline\hline
  \end{tabular}
\end{table}

Across most settings, we observe clear performance improvement on physical field prediction after introducing the stage 1 geometry-only pretraining. The results are consistent across different datasets and neural operators, showing that the pretraining in stage 1 potentially constructs a stronger geometry representation than directly feeding raw point clouds. Note that point clouds are discretized approximations of real geometries, while the latent representation (residing in function space since it reconstructs the occupancy field) offers a more accurate approximation, especially after being enriched on diverse geometry data. We also observe larger improvements for randomly sampled query points than for mesh-based queries. One potential reason is that mesh-based sampling is denser than uniform sampling, particularly near geometry boundaries so that baseline models readily capture critical geometric details, leaving less headroom for pretraining. This phenomon is most pronounced on the 3D Inductor dataset as uniform sampling in 3D space is especially inefficient.

\subsection{Ablation Study}
\label{subsec: ablation}

\paragraph{Number of geometry data.} As stage 1 physics-agnostic pretraining is decoupled from PDE solution labels, it is important study the impact of geometry data amount $|D\cup D'|$ on operator performance. Experiments are conducted on the Stress dataset where we pretrain VAEs on geometries sampled from four types of different void priors as shown in Table~\ref{table:abla1_vae}. We then compare the performance of GNOT and Transolver using the latent space from each VAE as shown in Table~\ref{table:abla1_result}. VAE1 is exposed to the same data $D$ as in stage 2, while VAE2 and VAE3 are trained on different settings of geometry data $D'$. It can be observed that the latent representation becomes stronger as more geometry data are added, even though some are sampled from different distributions. The improvement is less significant when the latent representation isn't augmented by extra geometry data (GNOT/Transolver+VAE1).

\begin{table}[!ht]
  \caption{Training metrics of stage 1 encoders using different number of geometry data on the stress dataset. Details of shape priors are discussed in Appendix~\ref{appendix:dataset}.}
  \label{table:abla1_vae}
  \centering
  \begin{tabular}{cccc}
    \hline\hline
    Stress data & VAE1 & VAE2 & VAE3 \\
    \hline\hline
    Ellipse $a=0.3$, $b=0.15$ & 1900 & 3800 & 1900 \\
    Ellipse $a=0.15$, $b=0.3$ & 0 & 0 & 1900 \\
    Rectangle $a=0.3$, $b=0.15$ & 0 & 0 & 1900 \\
    Rectangle $a=0.15$, $b=0.3$ & 0 & 0 & 1900 \\
    \hline
    IOU(\%) & 99.6 & 99.8 & 99.8 \\
    KL loss & 0.70 & 0.69 & 0.955 \\
    \hline\hline
  \end{tabular}
\end{table}
\begin{table}[!ht]
  \caption{Performance of stage 2 PDE learning using encoders learned on different number of geometry data.}
  \label{table:abla1_result}
  \centering
  \begin{tabular}{cccccc}
    \hline\hline
    Dataset & Query & \multicolumn{4}{c}{Relative L2 on normalized data ($\times 10^{-2}$)} \\
    \hline\hline
       &      & GNOT & G+VAE1 & G+VAE2 & G+VAE3 \\
    \cline{2-6}
    \multirow{6}{*}{\raisebox{0.7\height}{\shortstack{Stress\\Ellipse $a=0.3$, $b=0.15$\\1900}}} & Mesh & 9.8 & 9.9 & 9.7 & \textbf{9.0}  \\
    & Random & 10.3 & 9.1 & 8.6 & \textbf{8.3} \\
    \cline{2-6}
       &      & Trans & T+VAE1 & T+VAE2 & T+VAE3  \\
    \cline{2-6}   
    & Mesh   & 11.5 & 12.0 & 11.4 & \textbf{11.2} \\
    & Random & 11.5 & 10.4 & 9.8 & \textbf{9.7} \\
    \hline\hline
  \end{tabular}
\end{table}
\paragraph{Effect of KL weight.} We investigate of effect of KL weight ($\lambda$ in Eq~\ref{eq:vae_loss}). We train three VAEs and one deterministic autoencoder using the 2400 ArifRans geometries (Table \ref{table:dataset}) and report the IOU and KL loss in Table~\ref{table:abla2_kl}. The prediction accuracy of GNOT and Transolver using the encoded latent space of the four autoencoders are listed in Table~\ref{table:abla2_result}. It can be found all encoders enhance the performance of neural operators, and $\lambda=0.001$ (used in main experiments) is not necessarily optimal. Although all of the four autoencoders achieve near 100\% reconstruction IOU, weaker regularization constraint on latent space tends to provide more effective representations, but the trend is not absolute.
\begin{table}[!ht]
  \caption{Training metrics of point cloud autoencoder on the AirfRans (near) dataset at different KL weights.}
  \label{table:abla2_kl}
  \centering
  \begin{tabular}{ccccc}
    \hline\hline
    KL weight & $0.01$ & $0.001$ & $0.0001$ & Deterministic \\
    \hline\hline
    IOU(\%) & 99.9 & 99.9 & 99.9  & 99.9 \\
    KL loss & 0.06 & 0.36 & 1.62  & N/A \\
    \hline\hline
  \end{tabular}
\end{table}
\begin{table}[!ht]
  \caption{Prediction error of neural operators using latent geometry representation from VAEs trained with different KL weights. Here AE stands for deterministic autoencoder.}
  \label{table:abla2_result}
  \centering
  \renewcommand{\arraystretch}{1.05}
  \begin{tabular}{ccccccc}
    \hline\hline
    Dataset & Query & \multicolumn{5}{c}{Relative L2 on normalized data ($\times 10^{-2}$)} \\
    \hline\hline
       &      & GNOT & +VAE($0.01$) & +VAE($0.001$) & +VAE($0.0001$) & +AE \\
    \cline{2-7}
    \multirow{6}{*}{\raisebox{0.7\height}{\shortstack{AirfRans (near)\\$D$: 800+200\\$D\cup D'$: 2400+600}}} & Mesh & 6.8 & 6.6 & 5.6 & \textbf{5.3} & 5.9\\
    & Random & 7.8 & 6.2 & 5.9 & \textbf{5.2} & 5.9\\
    \cline{2-7}
       &      & Trans & +VAE($0.01$) & +VAE($0.001$) & +VAE($0.0001$) & +AE  \\
    \cline{2-7}   
    & Mesh   & 13.4 & 13.2 & 12.7 & 12.3 & \textbf{11.3}\\
    & Random & 15.0 & 11.1 & 10.8 & 11.1 & \textbf{10.3}\\
    \hline\hline
  \end{tabular}
\end{table}
\paragraph{Alternative proxy tasks.} While occupancy fields are selected as the default proxy task in the main experiments, our proposed framework is plug-and-play with alternative proxy task choices. In CFD problems such as AirfRans, SDF and SV are widely considered informative because they explicitly encode the relationship between arbitrary spatial locations and the nearest boundary, which is the dominant geometric factor in many CFD problems~\citep{jessica2023finite}. Therefore, we pretrain stage 1 with SDF or SV as the proxy task and evaluate stage 2 GNOT pressure prediction accuracy on AirfRans as shown in Table~\ref{table:abla3_result}. When compared with direct supervised operator learning, the proposed two-stage approach remains effective in most scenarios, except for the SDF case on mesh-based physics field query where the pretraining stage offers no measurable gain. Notably, using SV as the proxy task yields the best physics field prediction accuracy. These results demonstrate both the effectiveness and breadth of our framework beyond occupancy fields. However, SDF or SV are not necessarily as informative for PDE problems beyond CFD, especially when volumetric properties from body mesh dominate over surface proximity. Therefore, we still believe that occupancy is a more general, intuitive, and task-agnostic proxy choice.

\begin{table}[!ht]
  \caption{GNOT performance on the AirfRans dataset: our two-stage framework with occupancy, SDF, and SV proxy tasks in pretraining vs. the standard operator learning results (also with SDF or SV included in inputs).}
  \label{table:abla3_result}
  \centering
  \renewcommand{\arraystretch}{1.05}
  \begin{tabular}{ccccccc}
    \hline\hline
    \multirow{2}{*}{Query} & \multicolumn{6}{c}{Relative L2 on normalized data ($\times 10^{-2}$)} \\
    \cline{2-7}
       & GNOT & G+SDF & G+SV & G+OCC\_VAE & G+SDF\_VAE & G+SV\_VAE \\
    \hline\hline
    AirfRans (Mesh) & 6.8 & 5.2 & 5.4 & 5.6 & 5.3 & \textbf{4.8} \\
    AirfRans (Random) & 7.8 & 6.5 & 5.1 & 5.9 & 5.4 & \textbf{4.9} \\
    \hline\hline
  \end{tabular}
\end{table}


%% file: base/6_conclusion.tex
\section{Conclusion and Limitation}
\label{sec:conclusion}

The paper explores the feasibility of leveraging the abundant geometry data resource to augment the performance of neural operators on scarce physics labels. Specifically, we introduce a two-stage training work and propose to perform physics-agnostic pretraining on rich geometry data in stage 1 via an occupancy field reconstruction proxy task. The latent representation produced from the stage 1 encoder is then integrated to transformer-based neural operators for stage 2 physical field prediction learning. Incorporation of stage 1 pretraining yields consistent accuracy gains across different datasets and neural operator backbones. The framework is also adaptable to different proxy tasks and operator architectures. The results highlight the importance of properly selecting and preprocessing the representation of geometry inputs to neural operators.


The proposed physics-agnostic pretraining also possesses a few limitations. First, the effectiveness of occupancy reconstruction with multiple interacting geometries remains unexplored. Such settings may require replacing the BCE objective with multi-class cross entropy loss. Second, the default choice of occupancy is intuitive, and a systematic study is needed to help select from various alternative proxies including occupancy fields, SDF, SV, Gaussian density fields, etc., based on the governing physics, input geometry format, and proxy task accessibility. Lastly, integration between stages could be deepened beyond simply feeding Stage 1 latents. Potential strategies include joint or sequential fine-tuning with adjusted learning rates, hybrid losses that couple reconstruction and PDE supervision, or adapter-based conditioning.



%% file: base/appendix.tex
\newpage

\appendix

\section{LLM Usage}
\label{appendix:LLM}
LLM is only used on grammar polishing for this paper.

\section{Autoencoder Latent Space as a Chart on the Moduli Space of Geometric Structures}
\label{appendix:principle}

\paragraph{Designing Principles.} A central goal in shape analysis and geometry processing is to parametrize the space of geometric objects in a way that captures their essential structure while discarding irrelevant variations such as orientation, sampling density, or point ordering. This leads naturally to the idea of organizing geometric data—such as point clouds—into equivalence classes, where two point clouds are considered equivalent if they represent the same underlying shape up to transformations like rigid motion or reparameterization. The collection of these equivalence classes forms what is known in mathematics as a moduli space: a space whose points correspond to distinct geometric structures modulo some equivalence relation. In this framework, learning a meaningful representation of a point cloud becomes equivalent to coordinatizing the moduli space of shapes. An autoencoder offers a powerful data-driven approach to this problem: the encoder network acts as a learned coordinate chart, mapping each point cloud to a latent vector that serves as a representative for its equivalence class, while the decoder ensures that this latent code retains enough information to reconstruct the original geometry. Thus, the latent space can be interpreted as a learned, low-dimensional chart on the moduli space of point cloud geometries.

\paragraph{Formal Description.} Let $ \mathcal{X} \subset \mathbb{R}^{N \times d} $ denote the space of point clouds with $ N $ points in $ \mathbb{R}^d $, and define an equivalence relation $ \sim $ such that $ P_1 \sim P_2 $ if they represent the same underlying shape up to transformations from a group $ G $ such as permutation (automatically handled by transformer-based neural operators) and sampling/density variation (addressed in this work). Furthermore, rigid motion transformations can be easily incorporated into the proposed workflow with simple geometry data augmentation. The associated moduli space is the quotient space
\[
\mathcal{M} = \mathcal{X} / \sim,
\]
where each point $ [P] \in \mathcal{M} $ represents an equivalence class of point clouds corresponding to a single geometric structure.

An autoencoder consists of an encoder $ \mathcal{E}: \mathcal{X} \to \mathbb{R}^k $ and a decoder $ \mathcal{D}: \mathbb{R}^k \to \mathcal{X} $, trained to minimize a reconstruction loss
\[
\mathcal{L}(P) = \text{Dist}(P, \mathcal{D}(\mathcal{E}(P))),
\]
where $ \text{Dist} $ is a distance metric invariant under the group $ G $. Ideally, the encoder should be invariant under $ \sim $, i.e., $ P_1 \sim P_2 \Rightarrow \mathcal{E}(P_1) = \mathcal{E}(P_2) $, implying that it factors through the projection $ \pi: \mathcal{X} \to \mathcal{M} $. This gives the following commutative diagram:

\[
\begin{tikzcd}
\mathcal{X} \arrow[dr, "\pi"'] \arrow[rr, "E"] & & \mathbb{R}^k \\
& \mathcal{M} \arrow[ur, "\varphi"'] &
\end{tikzcd}
\]

\noindent where $ \mathcal{E} = \varphi \circ \pi $. While in practice $ \pi $ is not explicitly computed, and the encoder $ \mathcal{E} $ is trained end-to-end, this decomposition highlights the theoretical role of $ \mathcal{E} $: it should act as both a projection onto an equivalence class and a parametrization of that class. Therefore, the learned latent space $ \mathbb{R}^k $ can be interpreted as a local coordinate system on the moduli space $ \mathcal{M} $, and the encoder as a data-driven chart approximating this structure.

\section{Datasets}
\label{appendix:dataset}
In this section, we explain the details of the datasets including the governing equations, boundary conditions, material property, and geometry configurations. The original AirfRans dataset~\citep{bonnet2022airfrans} includes detailed airfoil boundary shape information for computing occupancy fields and generating new geometries. However, we choose to reproduce the Elasticity dataset~\citep{li2023fourier} as the detailed void geometry information is missing.

\paragraph{Stress.} We mimic the Elasticity dataset in~\citep{li2023fourier} and regenerate the geometries and stress fields in Fenics. We find the distribution of von Mises stress $\sigma_{vm}$ in a domain with an irregular void inside with a constant displacement vector at the top. We solve the following PDE:
\begin{equation}
\label{eqn: tension}
    \nabla\cdot\bm{\sigma}(u)=0
\end{equation}
where $u(x)$ is the displacement vector field, $\bm{\epsilon} = \frac{\partial u}{\partial x}$ is the strain tensor, and $\bm{\sigma} = \frac{\partial w(\bm{\epsilon})}{\partial \bm{\epsilon}}$ is the stress tensor. To obtain the energy $w(\bm{\epsilon})$ we need to start with Cauchy Green stretch tensor $\bm{C} = 2\bm{\epsilon} + \mathbb{I}$, and then get $w(\bm{\epsilon}) = C_1 (\mathrm{tr}(\bm{C}) - 3) + C_2(\frac{1}{2}(\mathrm{tr}(\bm{C})^2 - \mathrm{tr}(\bm{C})))$. Two energy density parameters are: $C_1 = 1.863 \times 10^5$ and $C_2 = 9.79 \times 10^3$. One can then plug this result back into Equation~\ref{eqn: tension} to get an equation in terms of $u(x)$. Note that although Equation~\ref{eqn: tension} is linear in terms of $\bm{\sigma}$, it is highly non-linear in terms of actual underlying field variable $u(x)$. After obtaining the displacement vector field, one can calculate stress tensor $\bm{\sigma}$, and then get von Mises stress field $\sigma_{vm}$ from components of $\bm{\sigma}$:
\begin{equation}
    \sigma_{vm} = \sqrt{\sigma_{xx}^2 - \sigma_{xx}\sigma_{yy}+\sigma_{yy}^2+3\sigma_{xy}^2}
\end{equation}

\begin{figure}
    \centering
    \includegraphics[width=0.9\linewidth]{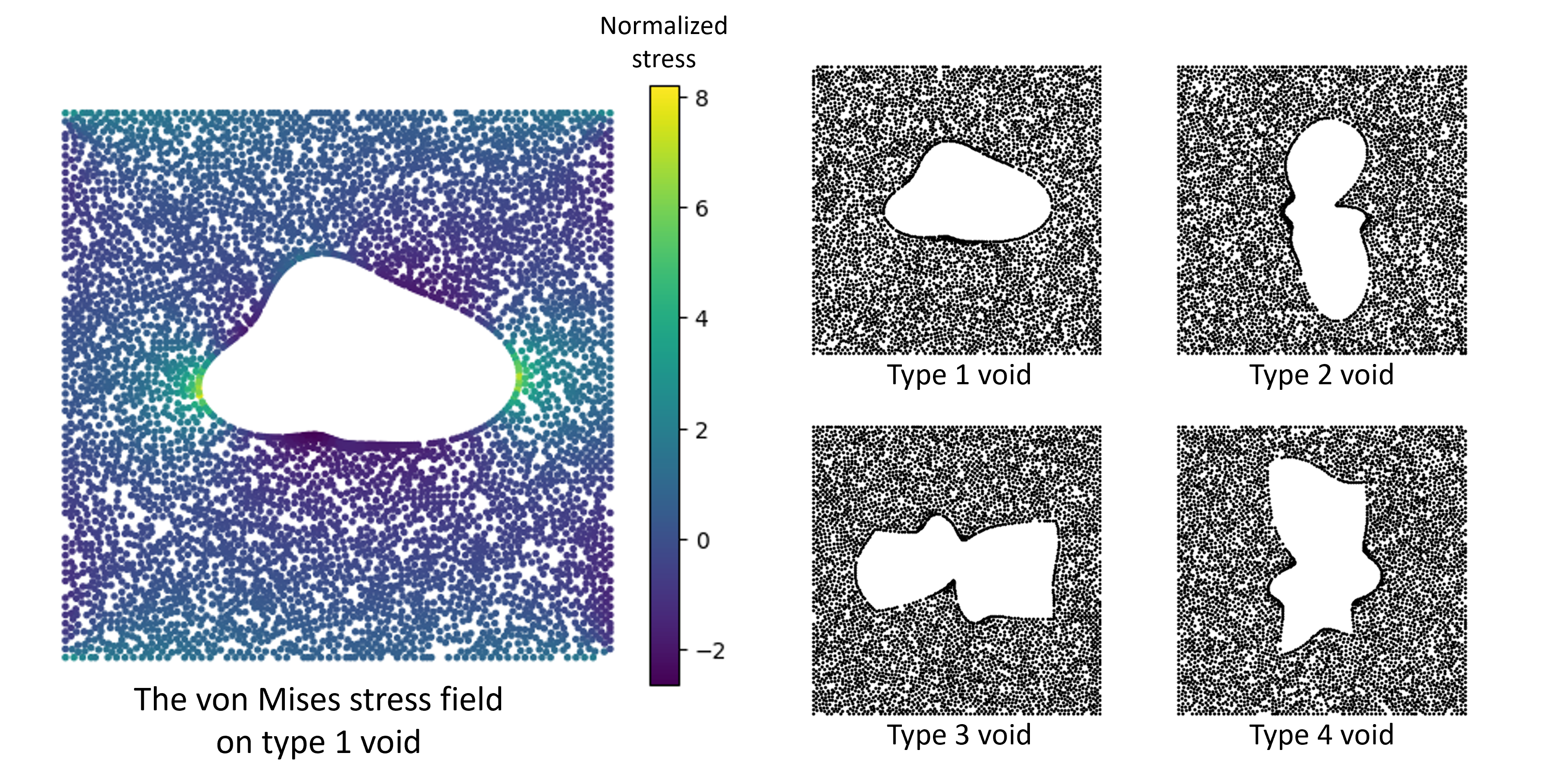}
    \caption{Examples of physics data and geometry data from the Stress dataset. Neural Operators only learn from von Mises stress fields on type 1 void, while observing the geometry data from geometries generated from 4 types of different void priors.}
    \label{fig:stress}
\end{figure}

For the solution domain, we have a square domain of $[0, 1]\times[0,1]$ with an irregular hole at the center. This hole is described by an expression in polar coordinates: $r = r(\theta)$. In our setup, $r(\theta) = f(a,b,\theta) + \mathrm{GRF}(A, l, \theta)$. The first $f$ term represents the prior void shape. The second term represents a Gaussian random field with an amplitude of $A=0.05$ and correlation length of $l=0.6$. We generate geometries using 4 types of different void priors as listed in Table~\ref{table:abla1_vae}, where $a$ represents the horizontal axis/edge of ellipse/rectangle and $b$ represents the vertical axis/edge of ellipse/rectangle. Example of the solution $\sigma_{vm}$ and geometries from different void priors are shown in Figure~\ref{fig:stress}. We generate 1900 training data and 100 testing data (occupancy fields) for each type of void for geometry representation learning. Neural operators are trained with 1900 data and tested with 200 data (von Mises stress fields) from type 1 void only.

\paragraph{AirfRans (near).} We directly adopt the simulation data in AirfRans~\cite{bonnet2022airfrans} which solves the Reynolds-Averaged Navier-Stokes Equations for airfoils. Unlike the other three datasets, AirfRans (near) contains two global features: attack velocity and angle, which are directly concatenated to all queries. The original dataset consists of 800 training data and 200 testing data with both geometry and physics information. We train transformer-based neural operators to predict the pressure fields around the airfoils in a truncated domain $[-0.6, 1.6]\times[-0.5, 0.5]$ as shown in Figure~\ref{fig:data_illustration}. To learn the geometry representation, we follow the same airfoil geometry parameterization (with a different random seed) in AirfRans and generate 2000 geometry only datasets (occupancy fields) with 1600 for training the VAE, and 400 for testing.

\begin{figure}
    \centering
    \includegraphics[width=0.9\linewidth]{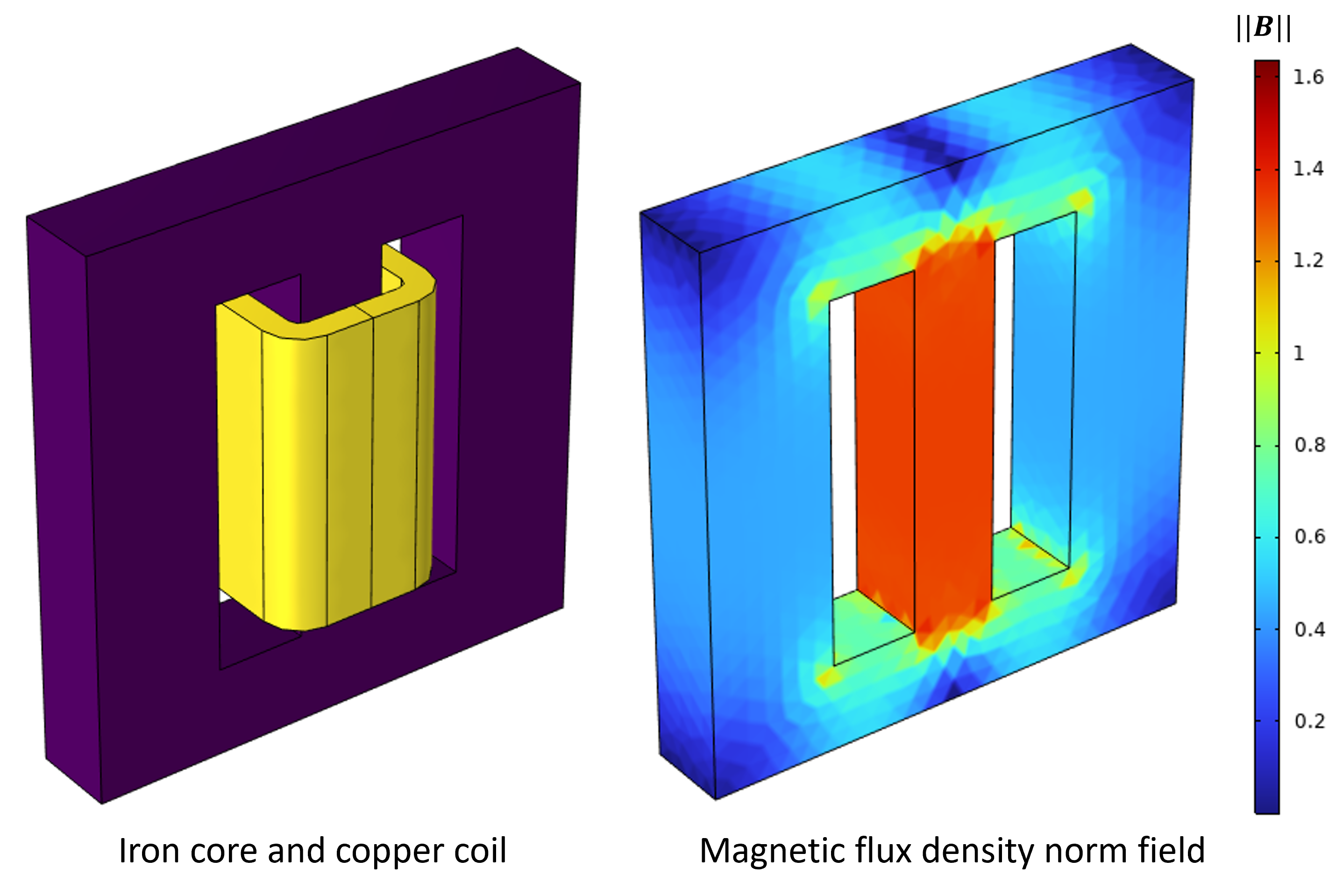}
    \caption{Example of a 3D inductor data. We generate the geometries (iron core in purple, coil in yellow) by parameterizing an EE type iron core, and calculate the corresponding magnetic flux density norm field (on the right).}
    \label{fig:inductor}
\end{figure}

\paragraph{Inductor 3D.} We calculate the magnetic flux density field $B$ around a 3D inductor by solving the Maxwell's equations (Eq~\ref{eq:maxwell}) in the frequency domain using Comsol.
\begin{equation}
\label{eq:maxwell}
\begin{split}
\nabla\cdot B&=0\\
\nabla\times H&=J+j\omega\epsilon E\\
\nabla\times E&=-j\omega\mu H
\end{split}
\end{equation}
where $H$ stands for magnetic field intensity, $E$ is the electric field, $J$ is the current density field, $\epsilon$ is material permittivity (which is assumed to be 1 everywhere), and $\mu$ is material permeability. Figure~\ref{fig:inductor} shows an example of the 3D inductor, consisting of an EE type iron core whose mid pillar is surrounded by a copper coil within a computation domain of $[-70,70]\times[-70,70]\times[-70,70]$ $mm^3$ (normalized for VAE and neural operator training). Magnetic insulation $n\times A$ is applied to the boundary of the computation domain, where $n$ stands for the unit normal and $A$ is the magnetic vector potential satisfying $B=\nabla\times A$. The copper coil contains 500 turns with a current density of $1e6$ $A/m^2$.

\begin{figure}
    \centering
    \includegraphics[width=0.6\linewidth]{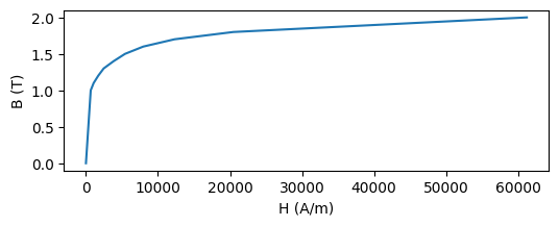}
    \caption{B-H curve of the iron core material.}
    \label{fig:bhcurve}
\end{figure}

The iron core follows a nonlinear magnetizing curve as shown in Figure~\ref{fig:bhcurve}. The neural operators are trained to predict the steady state magnitude of the magnetic flux density field $||B||$ at a high frequency of $f=\frac{\omega}{2\pi}=300$ kHz. The nonlinear material property curve and high excitation frequency make this 3D inductor dataset extremely challenging. We generate 3600 training data and 300 testing data by randomly selecting the geometry parameters of the iron core. We compute occupancy fields for all the data to train the VAE, while only 900 of the training data are simulated to train the neural operators as listed in Table~\ref{table:dataset}.

\begin{figure}
    \centering
    \includegraphics[width=0.8\linewidth]{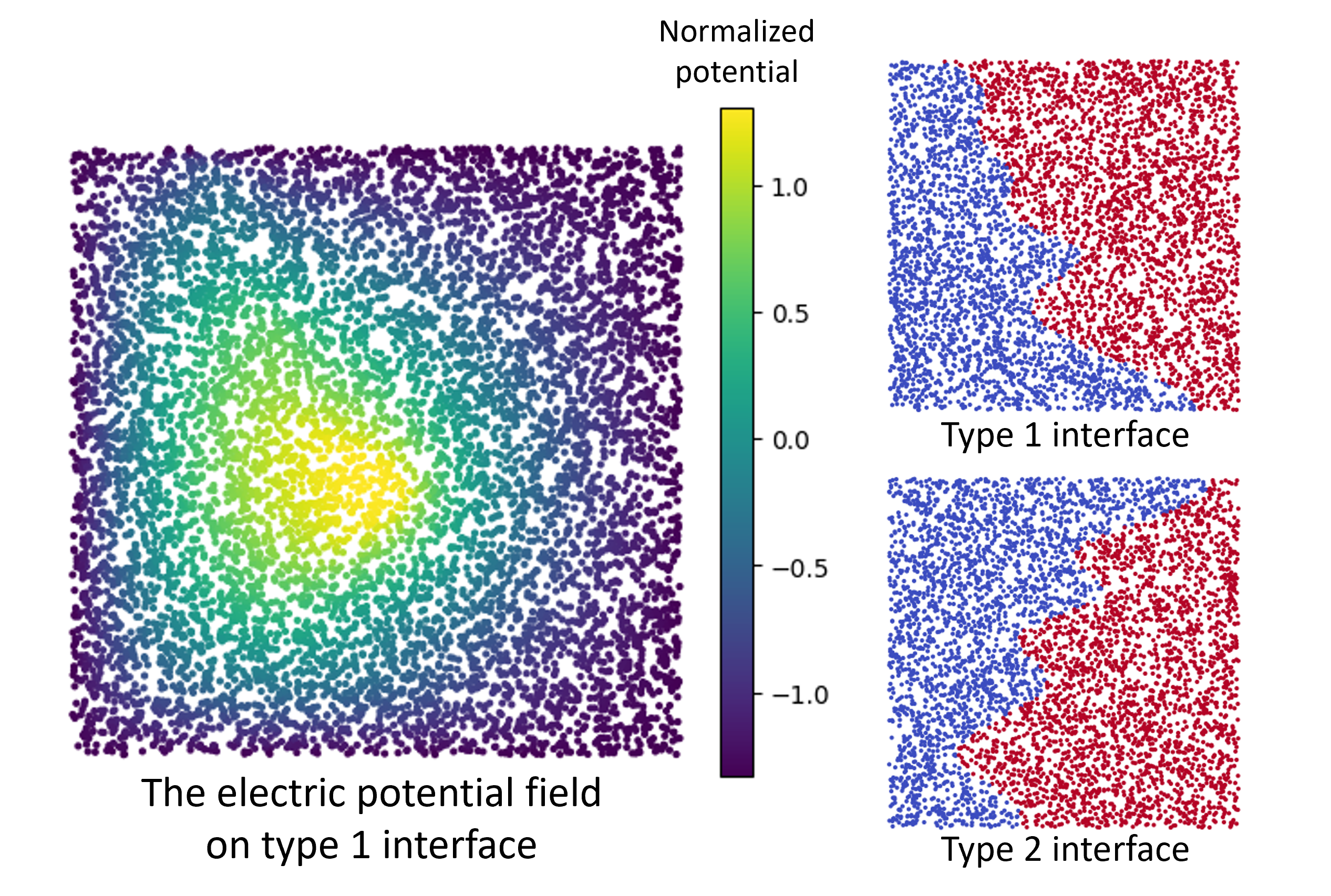}
    \caption{Examples of physics data and geometry data from the Electrostatics dataset. Neural Operators only learn from electric potential fields on type 1 interface, while observing geometry data generated from 2 types of different material interface priors.}
    \label{fig:elec}
\end{figure}

\paragraph{Electrostatics} An electrostatics dataset is built using Fenics where we solve for the electric potential $\phi(x)$ from the following equation:
\begin{equation}
    -\nabla\cdot(\epsilon(x)\nabla\phi(x)) = \rho(x)
\end{equation}
where $\epsilon(x)$ is the permittivity field, and $\rho(x)$ is the charge density field. Both of these fields are scalar fields. Here we follow the Heaviside-Lorentz units, so permittivity is unitless. In our dataset, the whole solution domain $\Omega$ is $[0,1] \times [0,1]$ is divided into two parts separated by a random interface. The random interface between these two parts is determined with $10$ random points. These $10$ points are interpolated with third order spline interpolation to produce a curve as shown in the right part of Figure~\ref{fig:elec}. The locations of the 10 random points are generated by adding a horizontal perturbation $\mathcal{U}(-0.2,0.2)$ to 2 types of prior locations: 10 equally distanced points from (0.3, 1) to (0.7, 0) and from (0.7, 1) to (0.3, 0). A charge density of $1$ and permittivity of $15$ is present in the blue part, while a charge density of $0$ and permittivity of $1$ is in the red part. The solution of this example case is shown in the left part of Figure~\ref{fig:elec}. We generate 3800 training data and 200 testing data (occupancy fields) for each type of interface for geometry representation learning. Neural operators are trained with 1800 data and tested with 200 data (potential fields) from type 1 interface only.

\section{Additional Experiments}
\label{apdx:addexp}
In this section, we present a few additional ablation studies to examine the effectiveness of the proposed latent geometry representation under different data, training, model selection, and hyperparameter settings.

\paragraph{Performance of UPT.} We also evaluate UPT on the Stress and Electrostatics dataset using our framework with the results shown in Table~\ref{table:aba7_result}. We implement a simplified version using transformer for encoder, decoder, and propagator, while omitting the message passing layer and inverse encoding decoding training to avoid extensive tuning. The results indicate that our framework can be smoothly adapted to UPT and improve its prediction accuracy on scarce physics labels. We notice that GNOT and Transolver (query at input layer) show superior performance compared to LNO and UPT (query at output layer). A likely reason is that UPT and LNO are tailored for latent temporal rollouts, whereas all four benchmarks here are stationary physics. We also expect further improvement in UPT with the complete architecture and training recipe.

\begin{table}[!ht]
  \caption{Prediction error of UPT (simplified implementation) on the Stress and Electrostatics datasets. The two-stage training framework outperforms direct supervised operator learning.}
  \label{table:aba7_result}
  \centering
  \renewcommand{\arraystretch}{1.05}
  \begin{tabular}{ccccc}
    \hline\hline
    \multirow{2}{*}{Model} & \multicolumn{4}{c}{Relative L2 on normalized data ($\times 10^{-2}$)} \\
    \cline{2-5}
       & Stress, Mesh & Stress, Random & Elec, Mesh & Elec, Random \\
    \hline\hline
    UPT & 26.3 & 20.8 & 11.6 & 12.0 \\
    \hline
    UPT+VAE & 12.8 & 10.6 & 4.9 & 4.8 \\
    \hline\hline
  \end{tabular}
\end{table}

\paragraph{Effect of physics data size.} We investigate the effectiveness of the latent geometry representation under different sizes of physics datasets. We expand the physics training data to 3800 for the Stress dataset ($\sigma_{vm}$ field on type 1 void) and the Electrostatics dataset ($\phi$ field on type 1 interface), while using the same VAE as in the main experiments (VAE3 in Table~\ref{table:abla1_vae}). Table~\ref{table:aba4_result} shows the performance of GNOT trained with different amount of physics data. For both datasets, we observe significant reduction in relative error for GNOT and GNOT+VAE when physics datasets are expanded. The performance improvement from replacing geometry representation is consistent across different sizes of physics data. Interestingly, the VAE used for the Stress dataset only observes 1900 random geometries generated from type 1 void prior, but still enhances the performance of GNOT when trained with 3800 physics data.

\begin{table}[!ht]
  \caption{Prediction error of GNOT (with and without stage 1), trained with different number of physics data. We use the same VAE as in the main experiments.}
  \label{table:aba4_result}
  \centering
  \renewcommand{\arraystretch}{1.05}
  \begin{tabular}{cccccc}
    \hline\hline
    \multirow{2}{*}{Dataset} & \multirow{2}{*}{Num of physics data} & \multicolumn{4}{c}{Relative L2 on normalized data ($\times 10^{-2}$)} \\
    \cline{3-6}
       &      & \multicolumn{2}{c}{GNOT} & \multicolumn{2}{c}{GNOT+VAE} \\
    \hline\hline
    \multicolumn{2}{c}{Query method} & Mesh & Random & Mesh & Random \\
    \hline
    \multirow{3}{*}{Stress} & 950 & 18.4 & 17.6 & 15.4 & 12.9 \\
    & 1900 & 9.8 & 10.3 & 9.0 & 8.3 \\
    & 3800 & 7.5 & 7.2 & 7.1 & 5.7 \\
    \hline
    \multirow{3}{*}{Electrostatics} & 900 & 6.4 & 6.7 & 5.1 & 5.3 \\
    & 1800 & 4.2 & 4.6 & 3.3 & 3.4 \\
    & 3800 & 3.1 & 3.3 & 2.3 & 2.3 \\
    \hline\hline
  \end{tabular}
\end{table}

\paragraph{Effect of batch size.} \label{para:bs} For efficient training, a large batch size of 100 is employed in the main experiments (including error estimation) and ablation studies. Additionally, we assess the proposed framework’s performance under a conventional batch size of 32. Table~\ref{table:aba5_result} presents the results of neural operators trained with identical settings as the main experiments~\ref{sec:exp}, except for the smaller batch size. A reduction in relative error can be observed for all experiments when trained at a smaller batch size. Meanwhile, the latent geometry representation improves the performance of neural operator in most experiments. However, its impact is minimal in four of the comparisons: GNOT on Stress with mesh query, GNOT on Inductor (3D) with both queries, and Transolver on Inductor (3D) with mesh query. We also observe performance drop when training Transolver on the Stress dataset. A potential explanation is that simply replacing Transolver's first layer with a linear cross-attention layer may be insufficient to fully leverage the latent geometry representation.

\begin{table}[!ht]
  \caption{Comparison of neural operator performance with batch size of 32. All other model/training hyperparameters remain the same as in the main experiments. Experiments demonstrating significant performance enhancement are highlighted in bolded fonts.}
  \label{table:aba5_result}
  \centering
  \renewcommand{\arraystretch}{1.05}
  \begin{tabular}{cccccccc}
    \hline\hline
    \multirow{2}{*}{Dataset} & \multirow{2}{*}{Query} & \multicolumn{6}{c}{Relative L2 on normalized data ($\times 10^{-2}$)} \\
    \cline{3-8}
       &      & GNOT & G+VAE & Trans & T+VAE & LNO & L+VAE \\
    \hline\hline
    \multicolumn{2}{c}{Number of parameters} & \multicolumn{2}{c}{1.7-1.8 M} & \multicolumn{2}{c}{1.7-1.8 M} & \multicolumn{2}{c}{1.8-1.9 M}\\
    \hline
    \multirow{2}{*}{Stress} & Mesh & 6.8 & 6.7 & 6.9 & 7.6 & 13.1 & \textbf{9.3} \\
    & Random & 6.1 & \textbf{5.5} & 6.2 & 6.7 & 14.4 & \textbf{6.5} \\
    \hline
    \multirow{2}{*}{AirfRans} & Mesh & 4.1 & \textbf{3.6} & 7.9 & \textbf{6.3} & 20.6 & \textbf{18.2} \\
    & Random & 5.6 & \textbf{3.6} & 8.9 & \textbf{6.4} & 16.9 & \textbf{5.8} \\
    \hline
    \multirow{2}{*}{Inductor (3D)} & Mesh & 4.8 & 4.9 & 5.6 & 5.6 & 10.7 & \textbf{7.5} \\
    & Random & 10.7 & 10.6 & 12.2 & \textbf{11.7} & 15.3 & \textbf{11.9} \\
    \hline
    \multirow{2}{*}{Electrostatics} & Mesh & 3.0 & \textbf{2.1} & 3.2 & \textbf{1.8} & 7.8 & \textbf{2.8} \\
    & Random & 3.3 & \textbf{2.3} & 3.8 & \textbf{1.8} & 6.9 & \textbf{2.8} \\
    \hline\hline
  \end{tabular}
\end{table}

\paragraph{Hidden dimension of neural operators.} Table~\ref{table:aba6_result} shows the results of neural operators with a hidden dimension of 256 and head number of 8, which drastically increases the number of learnable parameters. Larger neural operators achieve lower relative error compared to the main experiments in Table~\ref{table:main_exp}, with the learned geometry representation still outperforming point clouds in most cases. Minimal impact from the latent geometry representation is observed in two comparisons: GNOT on Stress with mesh query, and GNOT on Inductor (3D) with random query. Performance drop is observed in three cases: GNOT on Inductor (3D) with mesh query, Transolver on Stress with mesh query, and Transolver on Inductor (3D) with mesh query. Notably, this is the only scenario where applying the latent geometry representation leads to performance degradation in GNOT. A potential explanation is that the iron core only takes a relatively small occupancy volume in 3D, making it difficult for the VAE (trained with a BCE reconstruction loss) to capture the geometry accurately. The performance drop in Transolver has been discussed in \textbf{Effect of batch size}.

\begin{table}[!ht]
  \caption{Comparison of neural operator performance with hidden dimension of 256. All other model/training hyperparameters remain the same as in the main experiments. Experiments demonstrating significant performance enhancement are highlighted in bolded fonts.}
  \label{table:aba6_result}
  \centering
  \renewcommand{\arraystretch}{1.05}
  \begin{tabular}{cccccccc}
    \hline\hline
    \multirow{2}{*}{Dataset} & \multirow{2}{*}{Query} & \multicolumn{6}{c}{Relative L2 on normalized data ($\times 10^{-2}$)} \\
    \cline{3-8}
       &      & GNOT & G+VAE & Trans & T+VAE & LNO & L+VAE \\
    \hline\hline
    \multicolumn{2}{c}{Number of parameters} & \multicolumn{2}{c}{6.9 M} & \multicolumn{2}{c}{6.9 M} & \multicolumn{2}{c}{7.6 M}\\
    \hline
    \multirow{2}{*}{Stress} & Mesh & 7.6 & 7.7 & 7.7 & 8.6 & 22.4 & \textbf{10.2} \\
    & Random & 7.3 & \textbf{6.3} & 8.6 & \textbf{7.5} & 18.0 & \textbf{8.7} \\
    \hline
    \multirow{2}{*}{AirfRans} & Mesh & 6.0 & \textbf{4.1} & 9.2 & \textbf{8.5} & 25.3 & \textbf{13.9} \\
    & Random & 6.1 & \textbf{4.2} & 12.0 & \textbf{8.1} & 19.4 & \textbf{9.4} \\
    \hline
    \multirow{2}{*}{Inductor (3D)} & Mesh & 5.4 & 6.1 & 6.7 & 7.2  & 21.5 & \textbf{8.2} \\
    & Random & 11.4 & 11.3 & 13.4 & \textbf{12.0} & 15.9 & \textbf{12.1} \\
    \hline
    \multirow{2}{*}{Electrostatics} & Mesh & 3.4 & \textbf{2.5} & 4.3 & \textbf{2.2} & 10.2 & \textbf{3.1} \\
    & Random & 3.8 & \textbf{2.6} & 4.6 & \textbf{2.2} & 9.8 & \textbf{3.1} \\
    \hline\hline
  \end{tabular}
\end{table}

\paragraph{Size of latent tokens.} We ablate the stage 1 VAE latent configuration by comparing their effect on Stress dataset using GNOT in stage 2 as shown in Table~\ref{table:aba8_result}, where N*M in column headers indicates N tokens of dimension M. Results show that 256 latent tokens with 32 dimensions give the best performance across both mesh-based and random query strategies. Increasing token number or dimension doesn't yield consistent gains. Instead, the latent size appears to strike a balance between expressiveness and regularization.

\begin{table}[!ht]
  \caption{Comparison of stage 2 accuracy of GNOT on the Stress dataset using different sizes of stage 1 latent. Columns denote the latent configuration (\# token * latent dimension).}
  \label{table:aba8_result}
  \centering
  \renewcommand{\arraystretch}{1.05}
  \begin{tabular}{cccccccc}
    \hline\hline
    \multirow{2}{*}{Dataset} & \multicolumn{7}{c}{Relative L2 on normalized data ($\times 10^{-2}$)} \\
    \cline{2-8}
       & No VAE & 128*32 & 128*64 & 256*32 & 256*64 & 512*32 & 512*64 \\
    \hline\hline
    Stress, Mesh & 9.8 & 10.0 & 9.6 & \textbf{9.0} & 9.5 & \textbf{9.0} & 9.2 \\
    \hline
    Stress, Random & 10.3 & 8.4 & \textbf{8.3} & \textbf{8.3} & 8.6 & 8.8 & 8.6 \\
    \hline\hline
  \end{tabular}
\end{table}

\section{Compute Resources}
\label{appendix: computation}

We list the number of parameters and memory cost for training the VAEs and transformer-based neural operators in this section. Training time per epoch is not included as experiments are carried out on different GPUs (A100 or V100). To conduct experiments efficiently, we use a large batch size of 100 for all experiments. However, we also conduct extra experiments to examine the effectiveness of the proposed framework under a normal batch size of 32, as discussed in Appendix~\ref{apdx:addexp}. Other hyperparameters are listed in section~\ref{sec:exp}.

\begin{table}[!ht]
  \caption{Number of parameters and memory consumption for stage 1 VAE training and stage 2 neural operator (hidden dimension of 128) training. The difference among different experiments is negligible.}
  \label{table:resource}
  \centering
  \renewcommand{\arraystretch}{1.05}
  \begin{tabular}{cccccc}
    \hline\hline
    Metric & VAE Encoder & VAE Decoder & GNOT & Transolver & LNO \\
    \hline\hline
    Parameters (M) & 1.1 & 8.2 & 1.74 & 1.71 & 1.92 \\
    \hline
    Memory (GB) & \multicolumn{2}{c}{11.9} & 29.9 & 32.1 & 7.5 \\
    \hline\hline
  \end{tabular}
\end{table}

\section{Broader Impacts}
\label{appendix:impact}
This work introduces a two-stage physics-agnostic pretraining framework to enhance the performance of neural operators. We expect positive impact in research and industry fields where partial differential equations are solved to simulate the physical world. However, we don't envision any misusing of our methods or causing any negative social impact.